\documentclass[conference,a4paper]{IEEEtran}
\IEEEoverridecommandlockouts

\usepackage[hidelinks]{hyperref}
\usepackage[cmex10]{amsmath}
\usepackage{amssymb,amsfonts}
\usepackage{dblfloatfix}

\usepackage[ruled,vlined]{algorithm2e}
\usepackage{graphicx}
\graphicspath{{Figures/PDF/}{Figures/PNG/}}

\usepackage{booktabs}
\usepackage{siunitx}
\usepackage[numbers,compress]{natbib}
\usepackage{texnames}
\usepackage{bm,bbm}
\usepackage{orcidlink}

\usepackage{makecell}
\usepackage{hyphenat}

\usepackage{multirow}
\usepackage{cuted}
\usepackage{tocloft}
\usepackage[utf8]{inputenc}
\usepackage{setspace}

\usepackage{subcaption}
\usepackage{tabularx}
\usepackage{tikz}
\usepackage[parfill]{parskip}
\usepackage{setspace}
\usepackage{color,soul}

\makeatletter
\def\footnoterule{\kern-3\p@
  \hrule \@width 2in \kern 2.6\p@} 
\makeatother
\newcommand{\copyrightnotice}[1]{{%
  \renewcommand{\thefootnote}{}
  \footnotetext[0]{#1}%
}}

\begin{document}

\title{\uppercase{Vision-Language Agents for Interactive Forest Change Analysis}
}

\author{
\IEEEauthorblockN{
James Brock\orcidlink{0000-0002-1907-4224}, 
Ce Zhang\orcidlink{0000-0001-5100-3584}, 
Nantheera Anantrasirichai\orcidlink{0000-0002-2122-5781}
}
\IEEEauthorblockA{
\textit{University of Bristol, Beacon House, Queens Road, Bristol, United Kingdom} \\
Email: \{james.brock, ce.zhang, n.anantrasirichai\}@bristol.ac.uk
}
}

\maketitle

\copyrightnotice{The authors wish to acknowledge and thank the financial support of the UK Research and Innovation (UKRI) [Grant ref EP/Y030796/1] and the University of Bristol. For the purpose of open access, the author has applied a Creative Commons Attribution (CC BY) license to any Author Accepted Manuscript version arising.}

\begin{abstract}
    Modern forest monitoring workflows increasingly benefit from the growing availability of high-resolution satellite imagery and advances in deep learning. Two persistent challenges in this context are accurate pixel-level change detection and meaningful semantic change captioning for complex forest dynamics. While large language models (LLMs) are being adapted for interactive data exploration, their integration with vision-language models (VLMs) for remote sensing image change interpretation (RSICI) remains underexplored. To address this gap, we introduce an LLM-driven agent for integrated forest change analysis that supports natural language querying across multiple RSICI tasks. The system builds upon a multi-level change interpretation (MCI) vision-language backbone with LLM-based orchestration. To facilitate adaptation and evaluation in forest environments, we further introduce the Forest-Change dataset, which comprises bi-temporal satellite imagery, pixel-level change masks, and multi-granularity semantic change captions generated using a combination of human annotation and rule-based methods. Experiments show that the proposed system achieves mIoU and BLEU-4 scores of 67.10\% and 40.17\% on the Forest-Change dataset, and 88.13\% and 34.41\% on LEVIR-MCI-Trees, a tree-focused subset of the LEVIR-MCI benchmark for joint change detection and captioning (CDC). These results highlight the potential of interactive, LLM-driven RSICI systems to improve accessibility, interpretability, and efficiency of forest change analysis. All data and code are publicly available at https://github.com/JamesBrockUoB/ForestChat.
\end{abstract}


\section{Introduction}
Forests account for roughly 31\% of global land cover, providing habitat for up to 80\% of land-based species and delivering invaluable ecosystem services \cite{lines2022ai}. Global forest cover routinely fluctuates due to a myriad of pressures, such as industrial logging, agricultural expansion, wildfires, extreme weather events, disease, and pests \cite{hansen2013high}. Contemporary monitoring increasingly relies on automated remote sensing (RS) data collection and processing pipelines \cite{lines2022ai}. As access to forest-specific RS datasets has increased, a corresponding rise in deep learning and artificial intelligence methods for forest analysis has been observed \cite{yun2024status}. 

Recent work has explored integrating LLMs and VLMs for RSICI tasks \cite{yang2025change, liu2024change, irvin2025teochat}, motivated by moving beyond pixel-level change localisation toward semantic, queryable spatio-temporal understanding  \cite{li2024prospects}. Within forest RSICI, remote sensing change detection (RSCD) identifies where changes occur but provides limited semantic context on change drivers such as logging or agricultural expansion, whereas remote sensing image change captioning (RSICC) generates textual change descriptions but introduces challenges in visual grounding, temporal alignment, and linguistic precision \cite{liu2025remote, tao2025advancements}. VLMs that align pretrained visual encoders (e.g. CLIP, ViT) with LLMs have emerged to address these complementary tasks, enabling instruction-guided reasoning, zero-shot transfer, and interactive RS image querying \cite{lu2025vision, irvin2025teochat}. These developments mark a broader shift toward interactive, multi-modal, expert-in-the-loop workflows over fully automated pipelines \cite{li2024vision}. 

\begin{figure*}[ht]
    \centering
    \includegraphics[width=0.85\textwidth]{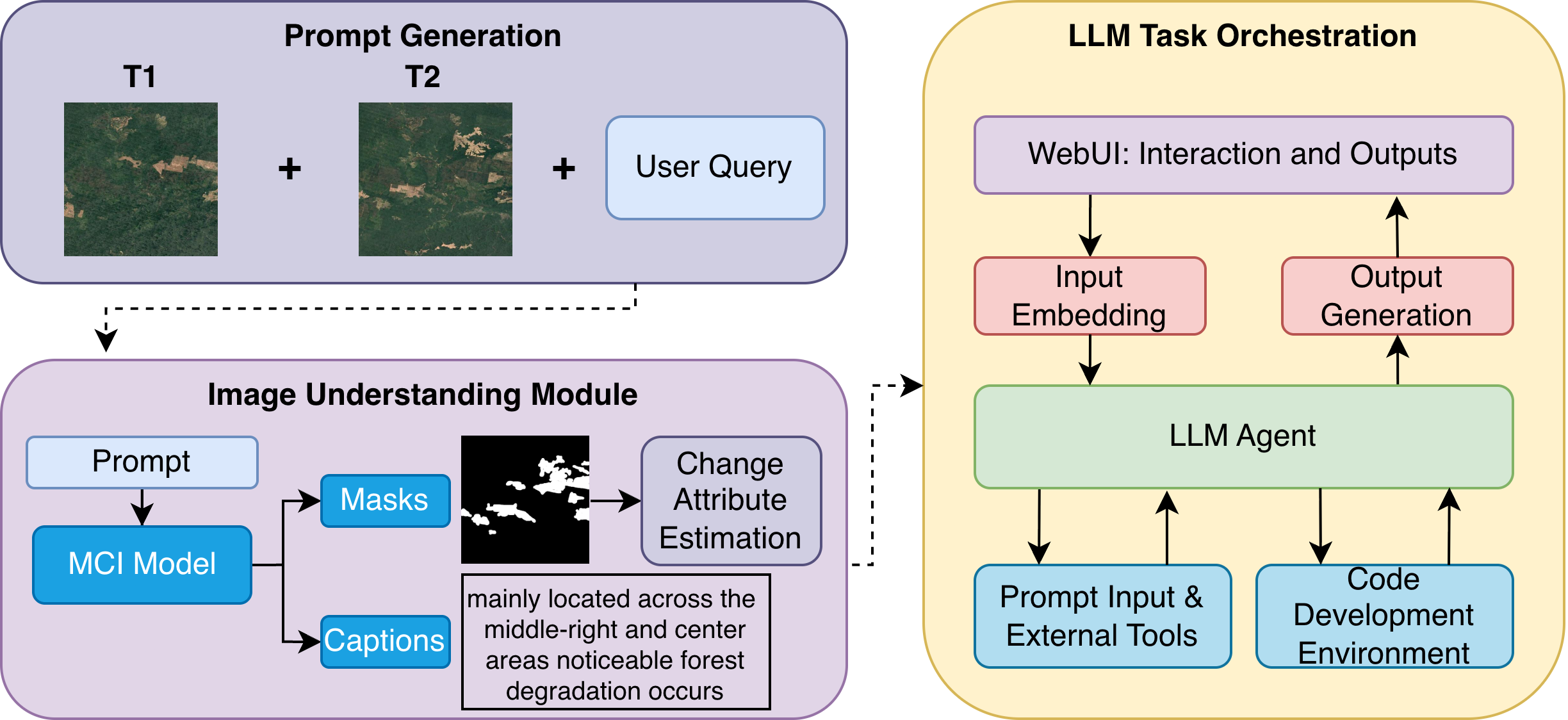}
    \caption[Graphical Abstract]{Overview of our proposed method. The system incorporates the MCI model and supplementary change analysis tools to operate as the system's eyes, with an LLM acting as its brain. The proposed Forest-Change dataset is used as a data foundation for training and adapting the MCI model for forest change analysis.}
    \label{fig:forest-chat-diagram}
\end{figure*}

More recently, LLM-driven RS vision-language agents have emerged as a natural extension of spatio-temporal VLMs, integrating perception, reasoning, and tool use within multi-step analytical workflows for Earth observation tasks \cite{liu2024change, irvin2025teochat}. Rather than producing fixed RSCD or RSICC outputs, these systems typically employ an LLM as a controller that interprets user intent, decomposes tasks, and selects or sequences specialised vision modules in response to intermediate results \cite{liu2025remote, guo2024remote}. Change-Agent \cite{liu2024change} exemplifies this paradigm for bi-temporal change analysis by coupling change-focused perception modules with LLM-guided reasoning to support interactive and interpretable assessments of temporal differences, while RS-Agent \cite{xu2024rs} demonstrates more general geospatial task orchestration through dynamic tool selection. Tree-GPT \cite{du2023tree} further illustrates the value of agentic design for forest scene analysis, combining LLM orchestration with vision modules, domain knowledge bases, and a web-based interface to support exploratory interpretation through powerful visualisation generation. However, Tree-GPT lacks explicit temporal reasoning capabilities, limiting its applicability for change-centric RS forest tasks. This, combined with the absence of any dataset supporting joint forest CDC, highlights a broader gap between emerging RS vision-language agents and the requirements of RSICI, where fine-grained spatial localisation, temporal alignment, and semantic explanation must be jointly addressed. Bridging this gap has motivated LLM-based agent frameworks explicitly designed for change analysis, such as TEOChat \cite{irvin2025teochat} and ChangeChat \cite{deng2025changechat}, which explore different trade-offs between modular orchestration and end-to-end instruction-tuned temporal reasoning.

In parallel, the emergence of foundation VLMs has motivated interest in zero-shot change interpretation. However, general-purpose multi-modal models such as GPT-4V \cite{openai2023} are limited in domain-specific RS tasks, performing poorly due to domain mismatch, weak spatial reasoning, and limited sensitivity to fine-grained or quantitative change \cite{zhang2024good}. This motivates the continued interest in supervised methods and dataset development for multi-task RSICI.

Building on the recent advances for RS vision-language agents, this work introduces an interactive agent tailored for forest change analysis (Figure \ref{fig:forest-chat-diagram}). The system integrates supervised dual CDC via the MCI model \cite{liu2024change}), supplemented by forest-specific analytical tools within a conversational LLM-driven interface, enabling the iterative exploration and interpretation of bi-temporal imagery. The proposed system enables a flexible, knowledge-driven exploration of forest change events, supporting both spatial localisation and semantic interpretation beyond the capabilities of conventional single-task models. To enable systematic evaluation, this paper further introduces the \textit{Forest-Change} dataset, the first dataset tailored explicitly for forest-based RSICI, providing aligned pixel-level change masks and semantic change captions. Extensive experiments benchmark our method against contemporary methods on both the proposed Forest-Change dataset and a forest-focused subset of LEVIR-MCI coined LEVIR-MCI-Trees \cite{liu2024change}.

The main contributions of this work are:

\begin{itemize}
    \item The introduction of an LLM-driven vision-language agent for interactive forest change analysis that, to the best of our knowledge, is the first to integrate joint supervised CDC via the MCI model, and forest-specific change analysis tools within a conversational interface.
    
    \item A purpose-built RSICI dataset specifically focused on forests called \textbf{Forest-Change}. It provides bi-temporal image pairs with pixel-level change masks and semantic change captions.

    \item A tree-focused subset of LEVIR-MCI named \textbf{LEVIR-MCI-Trees} is presented. It contains urban-focused change masks and forest-focused change captions, aimed at assessing generalisation from a larger, diverse data space to a smaller one with limited scene variety.
\end{itemize}

\section{Datasets}

\begin{figure}
    \centering
    \includegraphics[width=1\linewidth]{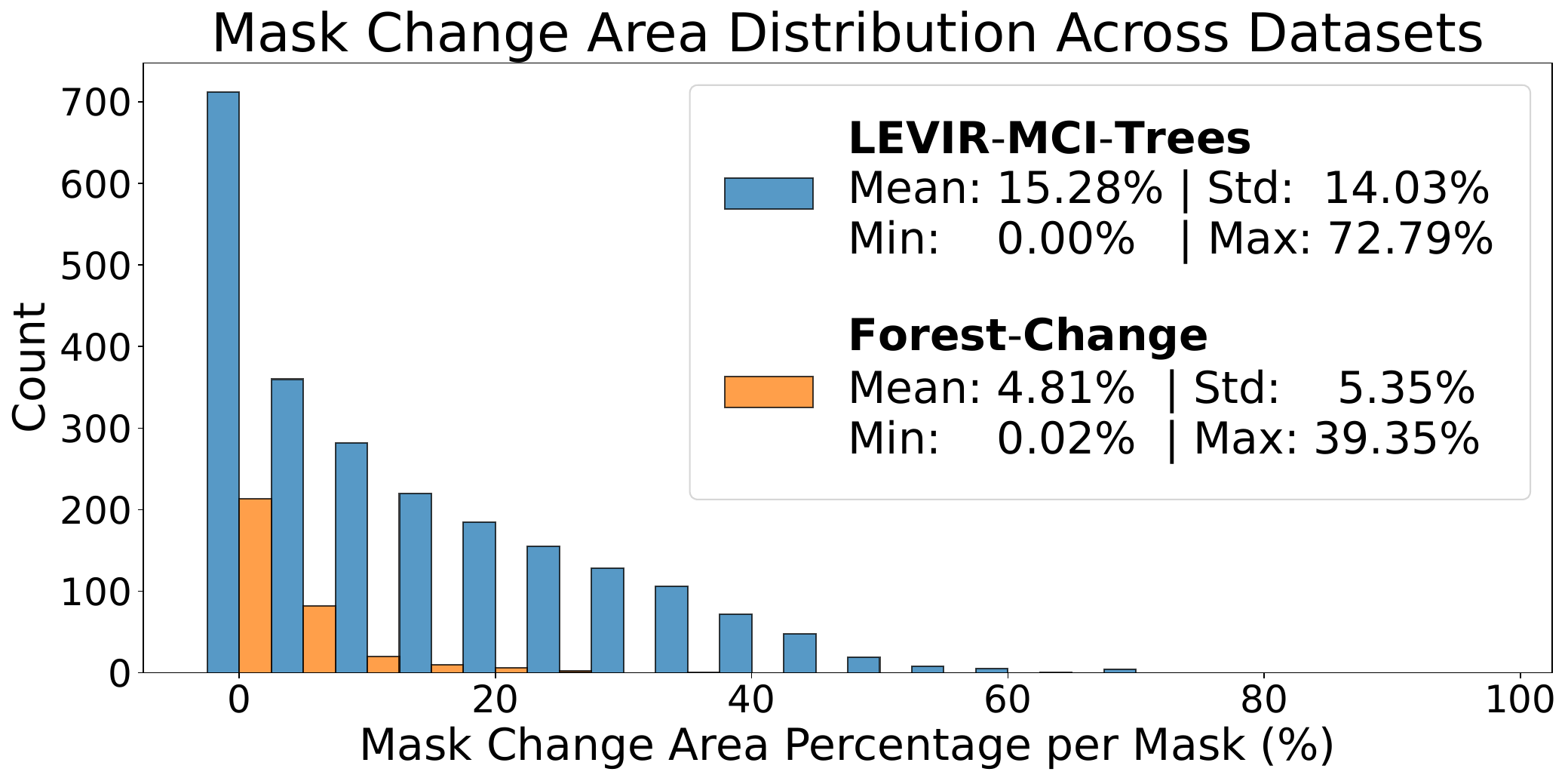}
    \caption[Dataset Segmentation Mask Statistics]{Summary statistics of change cover in segmentation masks for Forest-Change and LEVIR-MCI-Trees datasets.}
    \label{fig:dataset_mask_distribution}
\end{figure}

\begin{figure}
    \centering
    \includegraphics[width=1\linewidth]{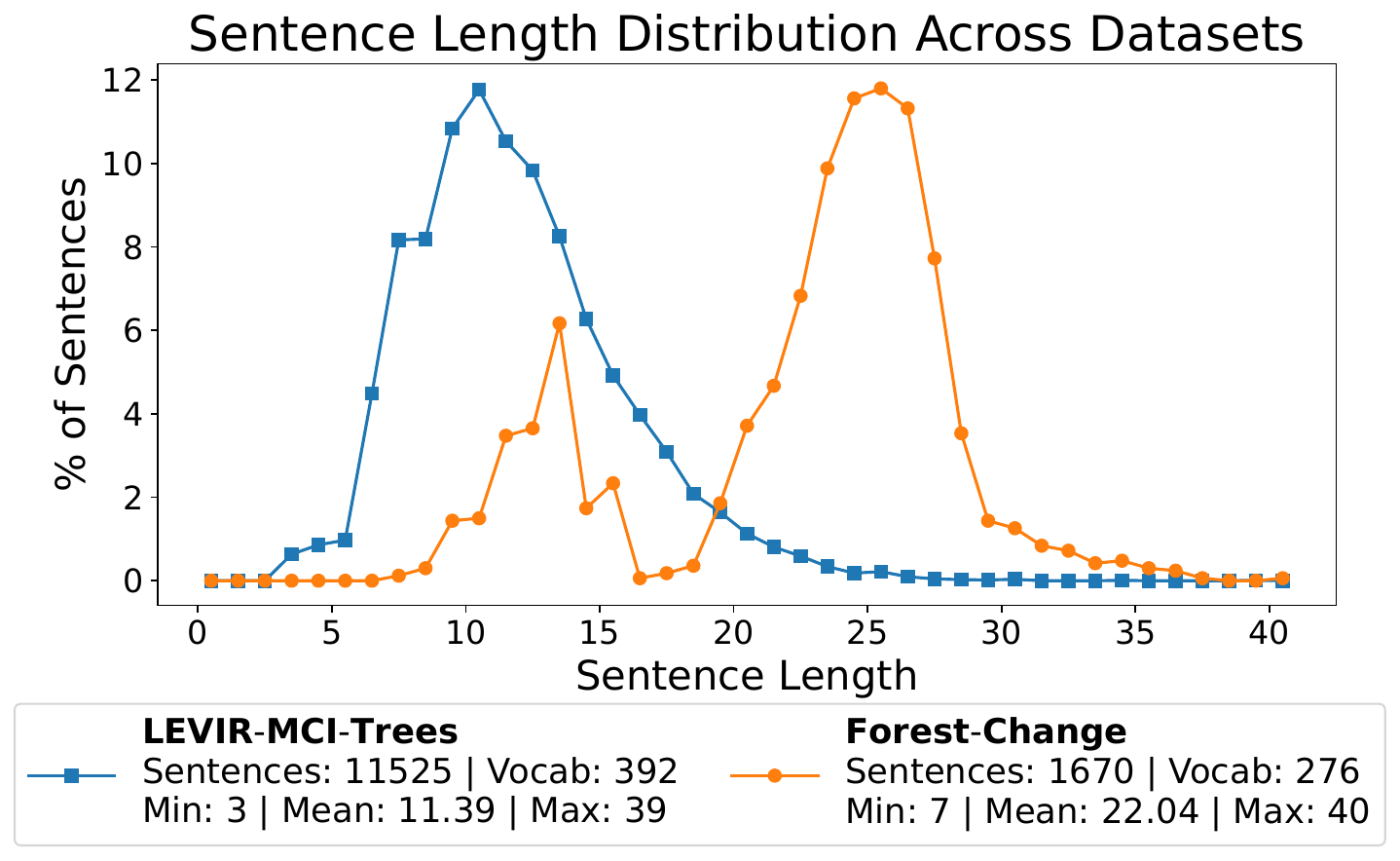}
    \caption[Dataset Sentence Statistics]{Summary statistics of  captions for Forest-Change and LEVIR-MCI-Trees datasets.}
    \label{fig:dataset_captions_distribution}
\end{figure}

This work utilises two datasets to support interactive forest change analysis. \textit{Forest-Change} is introduced here as the first forest-specific bi-temporal CDC dataset, while \textit{LEVIR-MCI-Trees} is a tree-focused subset of the LEVIR-MCI dataset \cite{liu2024change}, allowing evaluation of model generalisation from urban to forested scenes. Figures \ref{fig:dataset_mask_distribution} and \ref{fig:dataset_captions_distribution} provide graphical overviews of each dataset's mask and caption distributions.

\textbf{Forest-Change:} A forest-specific bi-temporal CDC dataset of tropical and subtropical RGB deforestation imagery from Hewarathna et al. \cite{hewarathna2024change}, sourced via Google Earth Engine \cite{gorelick2017google} at a medium spatial resolution of $\sim$30m/pixel with approximately one-year temporal intervals between image pairs. It contains 334 image pairs resized from 480$\times$480 pixels to 256$\times$256 pixels via bilinear interpolation, each with a binary change mask indicating forest loss. The relatively small dataset size reflects the limited number of source sites and filtering of cloud-obscured imagery. Most masks contain less than 5\% change, with a maximum of ~40\%. Captions were created through a custom-built application via a two-stage process: one human-authored description per pair, followed by four automatically generated captions derived from mask statistics (e.g., percentage of loss, patch size, spatial distribution) to minimise annotation fatigue and ensure adequate semantic context. Captions exhibit a bi-modal length distribution, resulting from the combination of rule-based generation and human annotations. The dataset is split into training, validation, and test sets of 270, 31, and 33 pairs, respectively. Images are pre-aligned and normalized, and masks binarised to indicate change (1) or no-change (0), supporting both pixel-level change detection and semantic captioning in forest change contexts. The dataset presents challenging characteristics, including subtle and spatially diverse forest loss patterns, small change regions, and high class imbalance, which together require models to capture fine-grained spatial and semantic detail.
 
\textbf{LEVIR-MCI-Trees:} Retains only examples from the LEVIR-MCI dataset \cite{liu2024change} - comprising high-resolution RGB Google Earth imagery (0.5m/pixel) with a variable 5-15 year temporal span - whose captions contain tree-related keywords (e.g., ``tree'', ``forest'', ``woodland''). This results in 2,305 examples distributed across training, validation, and test sets (1,518, 374, and 413 pairs, respectively) with 256$\times$256 pixels per image pair. Change masks indicate change for urban objects only (roads and buildings), ignoring wider scene changes. During evaluation, change masks are binarised into change/no change to align with Forest-Change. Each pair is accompanied by five captions from different interpretation perspectives, enabling evaluation of tree-related captioning while segmentation remains urban-focused. Mask coverage is higher than in Forest-Change (mean 15.28\%, max 72.79\%), and captions are generally shorter and more lexically diverse.

\section{Methodology}

Our work introduces an interactive VLM-based agent inspired by Change-Agent \cite{liu2024change} for forest change analysis. The system combines task orchestration via an LLM, a vision-language perception module, and a conversational interface to support multi-turn exploration of bi-temporal forest imagery. The LLM orchestrates a set of specialised tools, including the MCI model, to perform both pixel-level change detection and semantic change captioning. The MCI model aligns visual and language representations through a shared multi-task architecture built on a Siamese SegFormer backbone \cite{xie2021segformer} with \textbf{Bi-temporal Iterative Interaction (BI3)} layers. Multi-scale visual features support fine-grained boundary detection and high-level semantic reasoning, with low-level features refining change masks and high-level features driving caption generation. A convolutional-based projection layer facilitates the transition of visual features into text, with the resulting features fed into a transformer decoder to generate descriptive change captions. Both the change detection and change captioning branches employ cross entropy loss. During joint training, the two losses are normalised to the same order of magnitude to ensure each task contributes equally, with joint modelling of detection and captioning improving contextual understanding of change dynamics over single-task approaches.

These image understanding tools provide the foundation for the agent’s interactive capabilities, with additional downstream tools further supporting an expert-in-the-loop workflow for analysing and explaining forest change. The system is capable of producing and executing Python code to answer user queries. It is provided with few-shot prompt examples to help it decompose complex tasks into concrete steps to improve task execution. The agent’s conversational interface aims to reduce cognitive load and streamline workflows for domain experts. It facilitates efficient analysis of forest cover dynamics and enables iterative reasoning that combines automated perception with human insight. A high-level overview of the system is presented in Figure \ref{fig:forest-chat-diagram}.

\begin{table*}[t]
\centering
\caption{Change detection and change captioning performances on the test sets of LEVIR-MCI-Trees and Forest-Change datasets. Best results per dataset and metric are \textbf{bold}. Results are an average of three runs.}
\label{tab:combined-results}
\setlength{\tabcolsep}{4pt}

\begin{tabular}{c | c | c c c | c c c c c c c}
\toprule
Dataset & Model & mIoU & $IoU_{\textit{nc}}$ & $IoU_{\textit{c}}$ & B1 & B2 & B3 & B4 & METEOR & ROUGE$_L$ & CIDEr-D \\
\midrule
\multirow{3}{*}{LEVIR-MCI-Trees} 
 & BiFA \cite{zhang2024bifa} 
 & 87.54 & 95.63 & 79.45 
 & - & - & - & - & - & - & - \\

 & Change3D \cite{zhu2025change3d} 
 & 87.48 & 95.63 & 79.34 
 & 69.52 & 54.35 & 38.33 & 26.41 & 21.57 & 46.83 & 35.03 \\

 & Ours
 & \textbf{88.13} & \textbf{95.89} & \textbf{80.36} 
 & \textbf{75.25} & \textbf{60.90} & \textbf{46.21} & \textbf{34.41} & \textbf{23.32} & \textbf{49.34} & \textbf{48.69} \\

\cmidrule(l){1-12}
\multirow{3}{*}{Forest-Change} 
 & BiFA \cite{zhang2024bifa} 
 & \textbf{67.34} & 95.85 & \textbf{38.84} 
 & - & - & - & - & - & - & - \\

 & Change3D \cite{zhu2025change3d} 
 & 66.01 & 95.93 & 36.08 
 & 61.08 & 49.18 & 40.46 & 33.32 & 25.65 & 46.26 & 20.78 \\

 & Ours 
 & 67.10 & \textbf{96.12} & 38.07
 & \textbf{67.54} & \textbf{56.34} & \textbf{47.55} & \textbf{40.17} & \textbf{28.22} & \textbf{48.52} & \textbf{38.79} \\

\bottomrule
\end{tabular}
\end{table*}

\section{Experiments}
\textbf{Evaluation Metrics:}
Segmentation performance is evaluated using Mean Intersection over Union (mIoU) with per-class IoU to assess performance on the under-represented change class. Although LEVIR-MCI-Trees contains multiple semantic classes, evaluation is performed under a binary \textit{change (c)} / \textit{no change (nc)} setting to ensure consistency with Forest-Change. Change captioning performance is assessed using natural language generation metrics, including BLEU-1 to BLEU-4 (referred to as B1--B4) \cite{papineni2002bleu}, METEOR \cite{banerjee2005meteor}, ROUGE$_L$ \cite{lin2004rouge}, and CIDEr-D \cite{vedantam2015cider}, which together evaluate lexical overlap, fluency, recall of salient content, and semantic relevance.

\textbf{Experimental Setup:}
All models are implemented in PyTorch and trained via CPU on Isambard 3 \cite{green2025evaluation}. The maximum number of epochs is set to 100, with the backbone being trained until the sum of the mIoU and B4 scores does not improve for 10 consecutive epochs. Once the MCI model is initially trained, the backbone network is frozen, and training is continued on the best model for the change detection and change captioning branches separately. The Adam optimiser is utilised with an initial learning rate of 0.0001.

\begin{figure}
    \centering
    \includegraphics[width=1\linewidth]{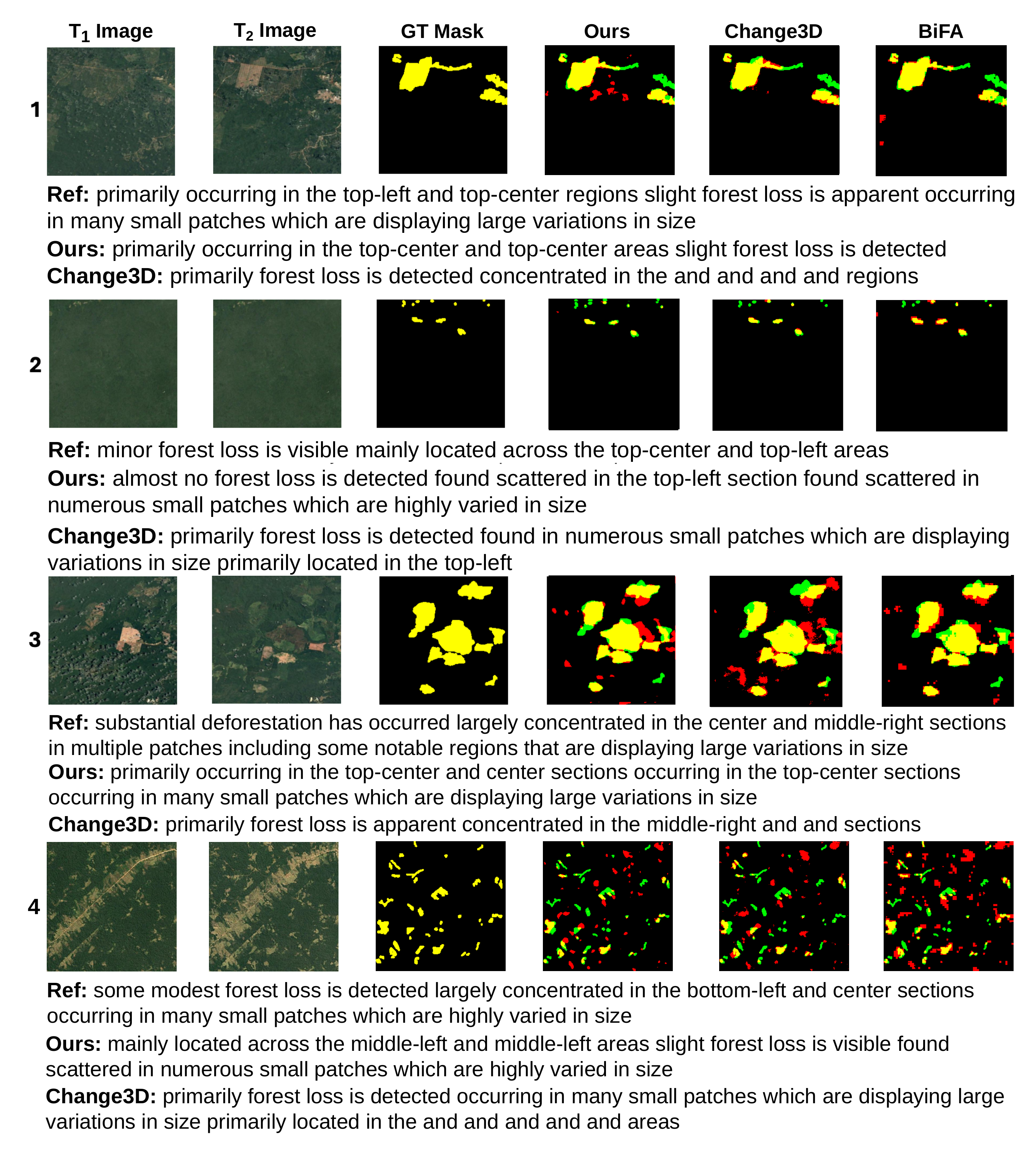}
    \caption[Qualitative Performance Examples]{A selection of qualitative comparison results between our approach and benchmarked models for CDC. Yellow indicates agreement with GT mask, red for false positives, green for false negatives. Best viewed when zoomed in.}
    \label{fig:qualitative_performance}
\end{figure}

\textbf{Result comparison:}
Due to the lack of methods that jointly perform CDC, our approach is evaluated against state-of-the-art models for each task separately on the LEVIR-MCI-Trees and Forest-Change datasets. While Change3D \cite{zhu2025change3d} supports both tasks, it is not implemented in a unified manner; we additionally select BiFA \cite{zhang2024bifa} for evaluating change detection. This experimental setup also enables a comparison across datasets of differing difficulty, contrasting a low-data forest scenario with a larger, more varied urban-focused benchmark.
Table \ref{tab:combined-results} summarises performance. On LEVIR-MCI-Trees, our approach achieves the best results for both the change detection and change captioning tasks, while on Forest-Change it ranks first for captioning and second for change detection, demonstrating strong generalisation across domains. Change detection is consistently lower for deforestation compared to buildings and roads, due to fewer training samples, pronounced class imbalances, small fragmented patches, and fuzzy or irregular boundaries; smaller deforestation patches are often partially detected, while larger regions are reliably captured. Captioning performance reflects the dataset-specific annotation styles: metrics such as B1, B2, and CIDEr-D are generally higher on LEVIR-MCI-Trees, whereas B3, B4, METEOR, and ROUGE$_L$ are higher on Forest-Change, highlighting the lexical richness of LEVIR-MCI-Trees and the rule-based prominence of Forest-Change captions. All models reliably capture change severity, but descriptions of location, patchiness, and geographic features can degrade for small, scattered regions. Qualitative analysis (see Figure \ref{fig:qualitative_performance} shows that captions are grammatically correct and include relevant descriptors, but models largely reproduce rule-based patterns from the training data, underutilising richer semantic language. Domain-aware approaches, such as vocabulary expansion \cite{gao2024ve} or domain-adaptive fine-tuning \cite{guo2022clip4idc}, could improve semantic richness and better represent geophysical and spatial characteristics.

\section{Conclusion}

This paper presents an interactive system for forest change analysis that combines a conversational interface with vision-language capabilities. An LLM agent integrates the vision-language capabilities and forest change analysis tooling to provide pixel-level change masks and semantic change captions. This enables users to explore bi-temporal forest imagery, characterise change patterns, assess forest health, and support ecological monitoring via natural language queries in a responsive and practical analytical workflow. We also introduce the Forest-Change and LEVIR-MCI-Trees datasets for joint CDC training and evaluation. Results show that the system reliably captures dominant change patterns and generalises from urban to forest contexts, though small, fragmented, or subtle ecological changes remains challenging.

\small
\bibliographystyle{IEEEtranN}
\bibliography{references}

@article{lu2025vision,
  title={Vision foundation models in remote sensing: A survey},
  author={Lu, Siqi and Guo, Junlin and Zimmer-Dauphinee, James R and Nieusma, Jordan M and Wang, Xiao and Wernke, Steven A and Huo, Yuankai and others},
  journal={IEEE Geoscience and Remote Sensing Magazine},
  year={2025},
  publisher={IEEE}
}

@article{xu2024rs,
  title={RS-Agent: Automating Remote Sensing Tasks through Intelligent Agent},
  author={Xu, Wenjia and Yu, Zijian and Mu, Boyang and Wei, Zhiwei and Zhang, Yuanben and Li, Guangzuo and Peng, Mugen},
  journal={arXiv preprint arXiv:2406.07089},
  year={2024}
}

@article{du2023tree,
  title={TREE-GPT: MODULAR LARGE LANGUAGE MODEL EXPERT SYSTEM FOR FOREST REMOTE SENSING IMAGE UNDERSTANDING AND INTERACTIVE ANALYSIS},
  author={Du, SQ and Tang, SJ and Wang, WX and Li, XM and Guo, RZ},
  journal={The International Archives of the Photogrammetry, Remote Sensing and Spatial Information Sciences},
  volume={48},
  pages={1729--1736},
  year={2023},
  publisher={Copernicus Publications G{\"o}ttingen, Germany}
}

@inproceedings{guo2022clip4idc,
  title={CLIP4IDC: CLIP for image difference captioning},
  author={Guo, Zixin and Wang, Tzu-Jui and Laaksonen, Jorma},
  booktitle={Proceedings of the 2nd Conference of the Asia-Pacific Chapter of the Association for Computational Linguistics and the 12th International Joint Conference on Natural Language Processing (Volume 2: Short Papers)},
  pages={33--42},
  year={2022}
}

@inproceedings{gao2024ve,
  title={VE-KD: Vocabulary-Expansion Knowledge-Distillation for Training Smaller Domain-Specific Language Models},
  author={Gao, Pengju and Yamasaki, Tomohiro and Imoto, Kazunori},
  booktitle={Findings of the Association for Computational Linguistics: EMNLP 2024},
  pages={15046--15059},
  year={2024}
}

@article{hewarathna2024change,
  title={Change detection for forest ecosystems using remote sensing images with Siamese attention U-Net},
  author={Hewarathna, Ashen Iranga and Hamlin, Luke and Charles, Joseph and Vigneshwaran, Palanisamy and George, Romiyal and Thuseethan, Selvarajah and Wimalasooriya, Chathrie and Shanmugam, Bharanidharan},
  journal={Technologies},
  volume={12},
  number={9},
  pages={160},
  year={2024},
  publisher={MDPI}
}

@article{zhang2024bifa,
  title={Bifa: Remote sensing image change detection with bitemporal feature alignment},
  author={Zhang, Haotian and Chen, Hao and Zhou, Chenyao and Chen, Keyan and Liu, Chenyang and Zou, Zhengxia and Shi, Zhenwei},
  journal={IEEE Transactions on Geoscience and Remote Sensing},
  volume={62},
  pages={1--17},
  year={2024},
  publisher={IEEE}
}

@inproceedings{zhu2025change3d,
  title={Change3D: Revisiting Change Detection and Captioning from A Video Modeling Perspective},
  author={Zhu, Duowang and Huang, Xiaohu and Huang, Haiyan and Zhou, Hao and Shao, Zhenfeng},
  booktitle={Proceedings of the Computer Vision and Pattern Recognition Conference},
  pages={24011--24022},
  year={2025}
}

@incollection{green2025evaluation,
  title={Evaluation of the Nvidia Grace Superchip in the HPE/Cray XD Isambard 3 supercomputer},
  author={Green, Thomas and Alam, Sadaf and McIntosh-Smith, Simon and Gilham, Richard and Wishart, William},
  booktitle={Proceedings of the Cray User Group},
  pages={93--102},
  year={2025}
}

@inproceedings{yang2025change,
  title={Change-UP: Advancing Visualization and Inference Capability for Multi-level Remote Sensing Change Interpretation},
  author={Yang, Mo and Chen, Luo and Zhou, Jiali},
  booktitle={Proceedings of the 33rd ACM International Conference on Multimedia},
  pages={15--24},
  year={2025}
}

@article{li2024vision,
  title={Vision-language models in remote sensing: Current progress and future trends},
  author={Li, Xiang and Wen, Congcong and Hu, Yuan and Yuan, Zhenghang and Zhu, Xiao Xiang},
  journal={IEEE Geoscience and Remote Sensing Magazine},
  volume={12},
  number={2},
  pages={32--66},
  year={2024},
  publisher={IEEE}
}

@article{liu2024change,
  title={Change-agent: Towards interactive comprehensive remote sensing change interpretation and analysis},
  author={Liu, Chenyang and Chen, Keyan and Zhang, Haotian and Qi, Zipeng and Zou, Zhengxia and Shi, Zhenwei},
  journal={IEEE Transactions on Geoscience and Remote Sensing},
  year={2024},
  publisher={IEEE}
}

@article{li2024prospects,
  title={Prospects for AI applications in forest protection: Technologies, challenges, and future developments},
  author={Li, Chengjun and Xiao, Mingkuan and Liu, Yunsheng},
  journal={Advances in Resources Research},
  volume={4},
  number={3},
  pages={362--380},
  year={2024},
  publisher={Resources Economics Research Board}
}

@article{xie2021segformer,
  title={SegFormer: Simple and efficient design for semantic segmentation with transformers},
  author={Xie, Enze and Wang, Wenhai and Yu, Zhiding and Anandkumar, Anima and Alvarez, Jose M and Luo, Ping},
  journal={Advances in neural information processing systems},
  volume={34},
  pages={12077--12090},
  year={2021}
}

@article{liu2025remote,
  title={Remote sensing spatiotemporal vision--language models: A comprehensive survey},
  author={Liu, Chenyang and Zhang, Jiafan and Chen, Keyan and Wang, Man and Zou, Zhengxia and Shi, Zhenwei},
  journal={IEEE Geoscience and Remote Sensing Magazine},
  year={2025},
  publisher={IEEE}
}

@article{tao2025advancements,
  title={Advancements in vision--language models for remote sensing: Datasets, capabilities, and enhancement techniques},
  author={Tao, Lijie and Zhang, Haokui and Jing, Haizhao and Liu, Yu and Yan, Dawei and Wei, Guoting and Xue, Xizhe},
  journal={Remote Sensing},
  volume={17},
  number={1},
  pages={162},
  year={2025},
  publisher={MDPI}
}

@inproceedings{irvin2025teochat,
  title={{TEOC}hat: A Large Vision-Language Assistant for Temporal Earth Observation Data},
  author={Jeremy Andrew Irvin and Emily Ruoyu Liu and Joyce C. Chen and Ines Dormoy and Jinyoung Kim and Samar Khanna and Zhuo Zheng and Stefano Ermon},
  booktitle={The Thirteenth International Conference on Learning Representations},
  year={2025},
  url={https://openreview.net/forum?id=pZz0nOroGv}
}

@inproceedings{guo2024remote,
  title={Remote sensing chatgpt: Solving remote sensing tasks with chatgpt and visual models},
  author={Guo, Haonan and Su, Xin and Wu, Chen and Du, Bo and Zhang, Liangpei and Li, Deren},
  booktitle={IGARSS 2024-2024 IEEE International Geoscience and Remote Sensing Symposium},
  pages={11474--11478},
  year={2024},
  organization={IEEE}
}

@inproceedings{deng2025changechat,
  title={Changechat: An interactive model for remote sensing change analysis via multimodal instruction tuning},
  author={Deng, Pei and Zhou, Wenqian and Wu, Hanlin},
  booktitle={ICASSP 2025-2025 IEEE International Conference on Acoustics, Speech and Signal Processing (ICASSP)},
  pages={1--5},
  year={2025},
  organization={IEEE}
}

@article{gorelick2017google,
  title={Google Earth Engine: Planetary-scale geospatial analysis for everyone},
  author={Gorelick, Noel and Hancher, Matt and Dixon, Mike and Ilyushchenko, Simon and Thau, David and Moore, Rebecca},
  journal={Remote sensing of Environment},
  volume={202},
  pages={18--27},
  year={2017},
  publisher={Elsevier}
}

@inproceedings{zhang2024good,
  title={Good at captioning bad at counting: Benchmarking gpt-4v on earth observation data},
  author={Zhang, Chenhui and Wang, Sherrie},
  booktitle={Proceedings of the IEEE/CVF Conference on Computer Vision and Pattern Recognition},
  pages={7839--7849},
  year={2024}
}

@article{openai2023,
  author={openai and Sally Applin and Gerardo Adesso and Rubaid Ashfaq and Max Bai and Matthew Brammer and Ethan Fecht and Andrew Goodman and Shelby Grossman and Matthew Groh and Hannah Rose Kirk and Seva Gunitsky and Yixing Huang and Lauren Kahn and Sangeet Kumar and Dani Madrid-Morales and Fabio Motoki and Aviv Ovadya and Uwe Peters and Maureen Robinson and Paul Röttger and Herman Wasserman and Alexa Wehsener and Leah Walker and Bertram Vidgen and Jianlong Zhu},
  title={GPT-4V(ision) System Card},
  year={2023},
  month={1},
  url={https://opal.latrobe.edu.au/articles/report/GPT-4V_ision_System_Card/25479208},
  doi={10.26181/25479208.v1}
}

@article{yun2024status,
  title={Status, advancements and prospects of deep learning methods applied in forest studies},
  author={Yun, Ting and Li, Jian and Ma, Lingfei and Zhou, Ji and Wang, Ruisheng and Eichhorn, Markus P and Zhang, Huaiqing},
  journal={International Journal of Applied Earth Observation and Geoinformation},
  volume={131},
  pages={103938},
  year={2024},
  publisher={Elsevier}
}

@inproceedings{lines2022ai,
  title={AI applications in forest monitoring need remote sensing benchmark datasets},
  author={Lines, Emily R and Allen, Matt and Cabo, Carlos and Calders, Kim and Debus, Amandine and Grieve, Stuart WD and Miltiadou, Milto and Noach, Adam and Owen, Harry JF and Puliti, Stefano},
  booktitle={2022 IEEE international conference on big data (big data)},
  pages={4528--4533},
  year={2022},
  organization={IEEE}
}

@article{hansen2013high,
  title={High-resolution global maps of 21st-century forest cover change},
  author={Hansen, Matthew C and Potapov, Peter V and Moore, Rebecca and Hancher, Matt and Turubanova, Svetlana A and Tyukavina, Alexandra and Thau, David and Stehman, Stephen V and Goetz, Scott J and Loveland, Thomas R and others},
  journal={science},
  volume={342},
  number={6160},
  pages={850--853},
  year={2013},
  publisher={American Association for the Advancement of Science}
}

@inproceedings{papineni2002bleu,
  title={Bleu: a method for automatic evaluation of machine translation},
  author={Papineni, Kishore and Roukos, Salim and Ward, Todd and Zhu, Wei-Jing},
  booktitle={Proceedings of the 40th annual meeting of the Association for Computational Linguistics},
  pages={311--318},
  year={2002}
}

@inproceedings{banerjee2005meteor,
  title={METEOR: An automatic metric for MT evaluation with improved correlation with human judgments},
  author={Banerjee, Satanjeev and Lavie, Alon},
  booktitle={Proceedings of the acl workshop on intrinsic and extrinsic evaluation measures for machine translation and/or summarization},
  pages={65--72},
  year={2005}
}

@inproceedings{lin2004rouge,
  title={Rouge: A package for automatic evaluation of summaries},
  author={Lin, Chin-Yew},
  booktitle={Text summarization branches out},
  pages={74--81},
  year={2004}
}

@inproceedings{vedantam2015cider,
  title={Cider: Consensus-based image description evaluation},
  author={Vedantam, Ramakrishna and Lawrence Zitnick, C and Parikh, Devi},
  booktitle={Proceedings of the IEEE conference on computer vision and pattern recognition},
  pages={4566--4575},
  year={2015}
}

\end{document}